\documentclass[11pt]{article}
\usepackage{fullpage}
\usepackage{comment}
\usepackage{amsmath}
\usepackage{amsthm}
\usepackage{amssymb}
\usepackage{multirow}
\usepackage{url}
\usepackage[final]{showkeys}	
\usepackage{tikz}
\usepackage{lineno}
\usepackage{color}
\usepackage{enumerate}
\usepackage{url}
\usepackage{amsmath, amsthm, amssymb, amsfonts, graphicx}
\graphicspath{{figs/}}
\usepackage{tikz}
\usepackage{enumerate}
\usepackage{cite}
\usepackage{clrscode}
\usepackage[labelfont=bf]{caption}
\usepackage{wrapfig}
\usepackage{subcaption}
\usepackage{bold-extra}
\usepackage[T1]{fontenc}
\usepgflibrary{shapes}
\usepackage[outline]{contour}
\contourlength{1pt} 

\usepackage{algorithmic}
\usepackage[ruled,vlined,commentsnumbered,titlenotnumbered]{algorithm2e}

\newtheorem{theorem}{Theorem}[section]
\newtheorem{lemma}[theorem]{Lemma}

\newtheorem{conjecture}{Conjecture}
\newtheorem{open}{Open Problem}

\newtheorem{definition}{Definition}[section]

\newtheorem{defn}[definition]{Definition}

\usepackage{hyperref}
\usepackage{cleveref} 

\def\be{\begin{enumerate}}

\def\ee{\end{enumerate}}
\def\bal{\begin{align*}}
\def\eal{\end{align*}}
\def\bi{\begin{itemize}}
\def\ei{\end{itemize}}
\def\bp{\begin{proof}}
\def\ep{\end{proof}}
\def\bl{\begin{lemma}}
\def\el{\end{lemma}}
\def\bt{\begin{thm}}
\def\et{\end{thm}}
\def\bc{\begin{cor}}
\def\ec{\end{cor}}
\def\bd{\begin{defn}}
\def\ed{\end{defn}}
\def\bprop{\begin{prop}}
\def\eprop{\end{prop}}

\usepackage[toc,page]{appendix}

{\makeatletter \gdef\fps@figure{!htbp}}

{\makeatletter
 \gdef\xxxmark{%
   \expandafter\ifx\csname @mpargs\endcsname\relax 
     \expandafter\ifx\csname @captype\endcsname\relax 
       \marginpar{xxx}
     \else
       xxx 
     \fi
   \else
     xxx 
   \fi}
 \gdef\xxx{\@ifnextchar[\xxx@lab\xxx@nolab}
 \long\gdef\xxx@lab[#1]#2{{\bf [\xxxmark #2 ---{\sc #1}]}}
 \long\gdef\xxx@nolab#1{{\bf [\xxxmark #1]}}
 \long\gdef\xxx@lab[#1]#2{}\long\gdef\xxx@nolab#1{}%
}

\makeatletter
\newcommand{\removelatexerror}{\let\@latex@error\@gobble}
\makeatother

\def \phs {\ensuremath{P_{\text{hs}}}}
\def \vhs {\ensuremath{V_{\text{hs}}}}

\begin{document}

\title{Recursed is not Recursive: A Jarring Result}
\author{
  Erik D. Demaine\thanks{Massachusetts Institute of Technology, \protect\url{edemaine@mit.edu}}
\and
  Justin Kopinsky\thanks{Work done while at Massachusetts Institute of Technology, \protect\url{jkopinsky@gmail.com}}
\and
  Jayson Lynch\thanks{Massachusetts Institute of Technology, \protect\url{jaysonl@mit.edu}}
}
\date{}
\maketitle

\begin{abstract}
Recursed is a 2D puzzle platform video game featuring \emph{treasure chests} that, when jumped into, instantiate a room that can later be exited (similar to function calls), optionally generating a \emph{jar} that returns back to that room (similar to continuations).
We prove that Recursed is RE-complete and thus undecidable (not recursive) by a reduction from the Post Correspondence Problem.
Our reduction is ``practical'': the reduction from PCP results in fully playable levels that abide by all constraints governing levels (including the $15 \times 20$ room size) designed for the main game.
Our reduction is also ``efficient'': a Turing machine can be simulated by a Recursed level whose size is linear in the encoding size of the Turing machine and whose solution length is polynomial in the running time of the Turing machine.
\end{abstract}

\def \pv {\textsc{Prove-Verify}}
\def \prv {\textsc{Prove}}
\def \ver {\textsc{Verify}}
\def \mem {\textsc{Mem}}
\def \gl {\textsc{Global-Lock}}
\def \poh {\textsc{Proof-of-Holding}}
\def \ow {\textsc{One-Way}}
\def \oto {\textsc{One-Time-Traversal}}
\def \dc {\textsc{Domino-Choice}}

\section{Introduction}

Recursed%
\footnote{All products, company names, brand names, trademarks, and sprites are properties of their respective owners. Sprites are used here under Fair Use for the educational purpose of illustrating mathematical theorems.}%
~\cite{recursed} is an indie puzzle platform video game by lone developer Portponky.
The game's main feature is having rooms contained within treasure chests, often recursively, inspired by functional programming; see Section~\ref{sec:rules} for details.

In this paper, we show that deciding whether a given Recursed level can be solved is RE-complete and thus undecidable (not recursive).%
\footnote{A brief recap on terminology: RE (Recursively Enumerable) is the class of decision problems whose ``yes'' instances are accepted by a Turing machine in finite time, but whose ``no'' instances may be indicated by the machine running for infinite time, while R (Recursive or Decidable) is the class of decision problems whose ``yes'' and ``no'' instances are accepted and rejected, respectively, by a Turing machine in finite time.  It is known that RE${}\subsetneqq{}$R; for example, the Halting Problem is in the difference RE${}\setminus{}$R.}
We thus positively settle a player's claim that Recursed is NP-hard~\cite{steamcurator} and another player's conjecture that it is undecidable~\cite{steamreview}.
Our proof is by a reduction from the Post Correspondence Problem (PCP); refer to Section~\ref{sec:proof} for a definition.
We use the properties of PCP that the constraints are locally checkable and that the resolution of choices proceeds in only one direction (adding more words/dominoes to the string).

RE-completeness of a video game requires some source of arbitrarily unbounded state in the game.
The unbounded state we use in Recursed stems only from the player's ability to generate instances of rooms arbitrarily deeply through normal use of the game's recursive chest mechanics (and by extension, its jar mechanics); see Section~\ref{sec:rules} for details.
Each of the finitely many rooms resulting from our reduction has constant size --- even fitting within the $15 \times 20$ size of standard Recursed rooms --- and contains only a constant number of objects and a constant amount of state.
Indeed, the Recursed levels generated by our reduction are ``practical'': they could, in principle, be solved by a human, provided they knew which PCP dominoes to place at each placement step.
Using the custom level feature of Recursed, we have built a fully playable custom level demonstrating the reduction applied to a simple 2-domino PCP instance, which is available for download \cite{xls2lua}.
\xxx{Once true: We have played it.}

Our reduction is also \emph{efficient}, meaning that it efficiently represents the execution of a Turing machine.
The Recursed level size is linear in the number $k$ of dominoes in the PCP instance, and the Recursed solution length is $O(L \log k)$ where $L$ is the number of symbols in a solution to the PCP instance.
Using the standard reduction from the Halting Problem to PCP \cite{sipser1996introduction}, the Recursed level size is linear in the encoding size $k$ of the Turing machine, and the Recursed solution length is $O(T S \log k) = O(T^2 \log k)$ where $T$ is the running time and $S$ is the space used by the Turing machine.
As a consequence, deciding whether a Recursed level can be solved in a polynomial number of steps is NP-complete, and deciding whether a Recursed level can be solved in an exponential number of steps is NEXPTIME-complete.

\paragraph{Related Work.}

The first RE-completeness/undecidability result for a video game was for another indie puzzle game, Braid \cite{braid}, designed by Jonathan Blow.
(The computational complexity of Blow's other puzzle game, The Witness, has also been studied, with NP-, $\Sigma_2$-, and PSPACE-completeness results for various aspects of the game \cite{abel2018witnesses}.)
To our knowledge, our result is the second RE-completeness/undecidability
result for a (real-world) single-player video game.

The Braid reduction \cite{braid} produces a Braid level of finite size.
The unbounded state it exploits comes from the game's ability to generate arbitrarily unbounded quantities of enemies and pack them into the same location, allowing the level to increment a counter arbitrarily high.
Enemies prevent the player from getting to a location, allowing the player to detect when the counter is zero.
In this way, the Braid reduction simulates a counter machine.
Because the reduction from Turing machine to counter machine \cite{minsky1961recursive} requires an exponential slowdown, the Braid reduction is not efficient: the resulting solution length is exponential in the running time of the Turing machine.
Also, because the items in Recursed all help rather than hinder the player's mobility, this type of approach cannot work for Recursed.

For two-player games, there is one undecidability result we are aware of:
Magic:\ The Gathering is RE-hard/undecidable even for two players
\cite{churchill2019magic}, via an efficient Turing machine simulation.
In fact, the players' moves are all forced, so this result is arguably about a zero-player simulation (but only the two-player game is ``real-world'').
An earlier RE-hardness/undecidability proof \cite{MtGTuring} simulated a counter machine, and thus was inefficient; it also required more players and a small tweak to the game rules.

Team multiplayer games are often RE-complete/undecidable even when the game's state is finite; the source of unboundedness is the hypothetical game strategies built in the players' heads \cite{GPCBook09}.
Recently, this technique has been applied to prove RE-completeness/undecidability of real-world team video games, including Team Fortress~2, Super Smash Brothers: Brawl, and Mario Kart \cite{coulombe2018cooperating}.
These reductions are naturally very different from Recursed, given the different source of unboundedness.




\paragraph{Roadmap.} Section~\ref{sec:rules} gives an overview of the mechanics of Recursed relevant to our construction. Section~\ref{sec:proof} presents a proof of the result. Section~\ref{sec:open} describes some open questions and conjectures regarding the complexity of subsets of Recursed.

\section{Game Rules}\label{sec:rules}

This section covers the rules of Recursed insofar as they are needed for our construction in Section~\ref{sec:proof}. For simplicity, we omit those objects and notions which we will not use.\footnote{Inexhaustively including water, acid, Ooblecks, cauldrons, paradoxes, glitches, and cloud walls.}

\subsection{Basic Player Actions}

We will call the player character Rico.
By default, Rico can \emph{run} horizontally and \emph{jump} or \emph{fall} vertically. Rico can jump up to a surface at most $3$ tiles higher than where they jumped from, but can fall arbitrarily far with no penalty. Rico can \emph{pick up} objects they are standing next to (see below for an enumeration). Rico can only carry one object at a time, and cannot pick up further objects until releasing the one held. While holding an object, Rico can \emph{drop} it, causing it to fall, or \emph{throw} it. Thrown objects travel in a perfectly vertical or horizontal trajectory until hitting a solid tile (or reaching the apex, if thrown upwards), at which point they fall until landing on a floor tile, or falling off the bottom of the screen. Note that other objects do not impede the horizontal trajectory of a thrown object, but, when falling, objects can land on blocks. If Rico is carrying an object, their jump height is lowered to at most $2$ blocks.

Typically, walls, floors, and ceilings are comprised of \emph{tiles}, which are immovable and impassable, by Rico or any object. At any given time, Rico will be situated in a room of size at most $15\times 20$ (though sometimes smaller), which is what is shown to the player. Typically, rooms will have borders consisting of solid tiles (counting towards the size). It is possible for some walls, floor, or ceiling to be missing. If Rico falls off the edge, they bounce back up a few blocks, but we will not have any missing floors in our construction.

A \emph{level} is comprised of a collection of rooms (see Section~\ref{sec:chests} for more details). Rico's goal is to reach a purple crystal, of which exactly one exists in some room of each level. 

There is one additional special environmental feature called a \emph{ledge} (see Figure~\ref{fig:objects}). Ledges are always oriented horizontally. Rico can jump \emph{upwards} through a ledge, but cannot by any means traverse back \emph{downwards} through a ledge. Thrown or dropped objects ignore ledges entirely in both directions. 

\subsection{Basic Objects}

\begin{figure}
\centering
\input{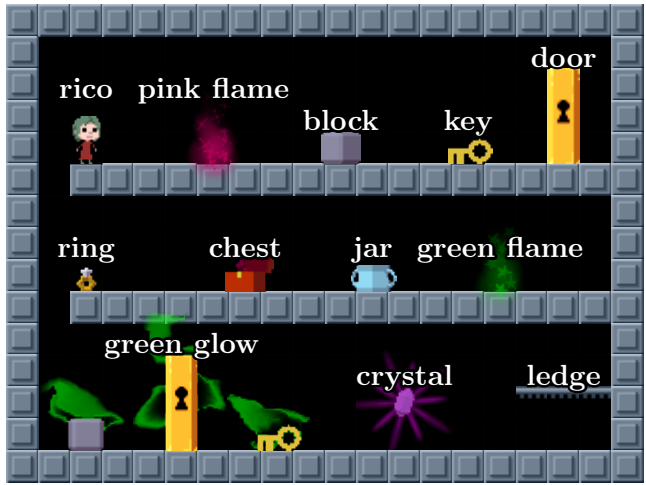}
\caption{A toy level containing a copy of each object used in our construction.}
\label{fig:objects}
\end{figure}

We depict all of the objects necessary for our construction in Figure~\ref{fig:objects}. A description of each follows.

\paragraph{Blocks} The primary use of blocks is for Rico to stand on and be able to jump higher. In particular, if Rico is standing on a block, they can jump to a height of $4$ above the ground ($3$ if carrying another object), which would be otherwise impossible. Furthermore, blocks can be stacked arbitrarily high, and Rico can ``climb'' stacks of blocks, so with $k$ blocks, Rico can reach a height of $k+3$. 

\paragraph{Keys and doors} Keys are carryable objects which open \emph{doors}. Doors are static objects in the level which occupy a space 3 tiles high by 1 tile wide, generally preventing traversal from one side to the other. A key can be carried directly to a door, or thrown at it. In either case, the key and the door both disappear, allowing Rico to traverse the space previously occupied by the door. There is only one type of key and one type of door, so any key in the game can open any door.

\paragraph{Rings} The only direct use of rings in-game is to trigger pre-scripted dialog when thrown against a wall. Therefore, they provide no particular use toward solving a level. Nevertheless, we will make use of rings in our construction as a generic object which specifically does nothing \emph{except} lower Rico's jump height to $2$ tiles when held. See Section~\ref{sec:poh}.

\subsection{Chests}
\label{sec:chests}
Chests are the primary ``gimmick'' of Recursed. They were designed to emulate function calls (in the programming sense) to some degree. In Recursed, chests do not contain objects, but rather entire rooms. Rico can jump \emph{into} chests, thereby entering the room contained therein. The room contained in a chest is an immutable property of the chest itself---a particular chest will always contain a particular room, according to the specification of that chest. Different chests can contain the same room.

When Rico enters a room via a chest, they appear at a pre-specified entry point\footnote{The entry point is a property of the room itself; Rico will appear at the same entry point regardless of which chest was used to enter the room.}, along with a \emph{pink flame}, which we refer to interchangeably as an \emph{exit}. The room will be generated freshly from its specification each time Rico enters a chest containing it \emph{regardless of whether Rico has previously visited and/or interacted with objects in the room}. This follows the function call intuition of each invocation entering the function at the beginning with no local state.

While in a room contained in a chest (which is most of the time), Rico can freely interact with any objects present \emph{including other chests}. Every room except for the initial room Rico begins the level in will necessary have a pink flame exit. If Rico returns to the exit of a room, they can choose to leave by interacting with the pink flame. In doing so, they will hop back out of the chest they initially came in, thus re-entering the ``parent'' room \emph{in the same state that it was when Rico jumped in the chest}. Note the asymmetry between entering and exiting chests. Again, this emulates the function call behavior of saving the local state of the parent function when returning from a child function. Because any future entries to the chest room will re-generate the room anew, any state that room had when Rico leaves is entirely forgotten.

Rico can of course recursively enter chests (hence the name), thereby saving a ``call history'' in a stack-like fashion. Rico can even enter a room via a chest contained in that same room, reminiscent of a recursive function calling itself. 

One final key property of chests is that when Rico jumps into a chest, or leaves via the pink flame, they can do so while carrying at most one object. Thus, provided that Rico can manage to get access to it, Rico can bring a block, a key, or another chest with them into or out of a chest. To demonstrate the impact of this ability, observe that on the one hand, if Rico carries, say, a block into a chest and then subsequently leaves the chest empty-handed, that block is \emph{lost forever}. On the other hand, if Rico enters a chest empty-handed, but manages to leave while carrying a block, the parent room now has a block that in effect did not previously exist. In particular, Rico can repeat the same sequence of actions any number of times to produce an \emph{unbounded} number of blocks in the parent room. 

\subsection{Green glow}
\label{sec:greenglow}

Some objects in the game have a \emph{green glow} (see Figure~\ref{fig:objects}). These objects violate the ``function call'' rules of chests described above, in that the state of a green glowing object is saved no matter when or where it is interacted with. The simplest example is a green glowing door, since it cannot be moved, but only open. If a green glowing door in a room is opened by any key, it will \emph{always} be open when Rico revisits that room, even if doing so by entering a chest and thus regenerating (the nonglowing parts of) the room. 

Movable objects, including blocks, keys, and chests can also glow green. In this case, if the object moves around the room it begins the level in, then whenever Rico revisits that room the location of the object will be remembered. This is the only property we will make use of in the construction, but we note for posterity that green glowing objects can be moved between rooms and this will be remembered as well. Green glowing chests have even more interesting properties, but we encourage the reader to play the game and discover those for themselves!

\subsection{Jars}
\label{sec:jars}
Jars are similar to chests in that they contain rooms, but unlike chests, they are designed to emulate \emph{continuations} (in the functional programming sense), rather than function calls. Jars can never be present in the initial state of a level. Rather, some rooms (other than the initial starting room) will have a \emph{green flame} exit in addition to the standard \emph{pink flame} exit (not to be confused with green glow above). The green flame can be located anywhere in the room and is independent of Rico's initial point of entry. There can even be more than one (though not in our construction). If Rico exits a room via a green flame exit, they will hop back out of the containing chest, just like by the pink flame, except they will now be carrying a newly created \emph{jar}. Note that Rico cannot be carrying any object when leaving via green flame exits in order to have space to carry the jar. 

Rico can carry the jar around just like any other object. If, subsequently, Rico enters a jar, they will re-enter the room containing the green flame the jar was created with, at the location of the green flame, with the room in the \emph{same} state that it was in when the jar was created, \emph{except} that the green flame itself is now gone, so no further jars can be created from the same place. Thus, any doors previously opened or objects previously moved or placed will be just where they were when Rico used the green flame. Importantly, when Rico later exits a room after entering it from a jar, they will reappear in the parent room just as if they had used a chest, and may even be carrying an object, but the jar will be destroyed. Thus, any particular jar can only be entered once.

\section{Main Result}
\label{sec:proof}

\begin{theorem}
Recursed is RE-complete.
\end{theorem}

Containment is straightforward: the game can obviously be simulated, given an initial state and sequence of player inputs. Thus, with a recursively enumerable Turing machine, one can enumerate every input string frame-by-frame and check whether any such string solves the level.

The hardness reduction is from the Post Correspondence Problem (PCP). Originally shown undecidable by Post in \cite{post1946variant}, we follow Sipser's description of the problem \cite{sipser1996introduction}. Given a set of dominoes $D_1, \dots, D_k$ each with a string $A_i = a_{i1}a_{i2}\dots a_{is_i}$ on the top half and a string $B_i = b_{i1}\dots b_{ir_i}$ on the bottom half, denoted $D_i = \langle A_i \mid B_i \rangle$. We are tasked with laying such dominoes next to each other (copying dominoes as much as necessary) such that the concatenation of the top halves equals the concatenation of the bottom halves. We will enforce that the first domino must be $D_0$ which will simplify the initial part of the construction. (Forcing the first domino to be of a specified type clearly does not make the problem decidable, since if it did there is a trivial nondeterministic decision algorithm which guesses the first domino and then calls the hypothesized decision oracle).

We will implement a (nondeterministic) algorithm to solve PCP in Recursed. The algorithm is as follows:

\begin{enumerate}
\item Nondeterministically choose a domino $D_i = \langle A_i \mid B_i \rangle$ to add to the solution, or stop and skip to~\ref{algsketch:pop}. \label{algsketch:start}
\item Push $A_i$ onto stack $S_A$ and $B_i$ onto stack $S_B$.
\item Return to~\ref{algsketch:start}.
\item Pop $S_A$ and $S_B$ and check whether the popped symbols are equal; REJECT if not. \label{algsketch:pop}
\item Repeat~\ref{algsketch:pop} until one stack empties, then check if the other stack is also empty; if yes, ACCEPT, else REJECT.
\end{enumerate}

\subsection{High-level overview}
\label{sec:highlevel} 
\begin{figure}
\centering
\input{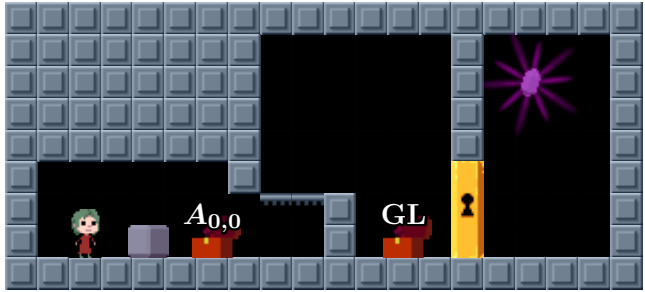}
\caption{The initial room.}
\label{fig:start}
\end{figure}

Rico initially spawns in a room next to a block and a chests, with another chest and a locked door past a ledge, shown in Figure~\ref{fig:start}. On the other side of the locked door is the goal crystal, but it is 6 tiles above the ground, one tile too high for Rico to jump to reach unaided\footnote{Rico can reach a jump height of 3 tiles and is 2 tiles tall, and so can reach a crystal 5 blocks high.}. The chest next to the door is a \gl{} chest, which, once unlocked, will provide the key to this very door. All Rico has to do is open the \gl{} gadget and get the block to the other side of the locked door! Of course, it will not be so easy...

The reduction is demonstrated with a fully playable level~\cite{xls2lua} for a PCP instance with $D_0 = \langle 01 \mid 0 \rangle$, $D_1 = \langle 0 \mid 10 \rangle$, whose (shortest) solution is of course $D_0D_1$. 

The high level structure of the construction is as follows: We will store $S_A$ in the ``call history'' of chests Rico has jumped into and we will store $S_B$ in a chain of jars. For each symbol $a_{ij} \in S_A$, there will be one corresponding room in our call history. For each symbol $b_{ij} \in S_B$, there will be one corresponding room in our chain of jars. Rico will need to carry the top level jar around pretty much all the time, unless changing the state of another gadget. The very last jar at the bottom of the chain representing $S_B$ will contain a single block\footnote{One might be forgiven for referring to this as the ``blockchain representation'' of $S_B$.} which, if retrieved in the crystal room, Rico can use to jump on and reach the crystal.

The intended solution path which Rico must follow is comprised of two phases: a ``Pushing'' phase, and a ``Checking'' phase. During the pushing phase, Rico will push symbols to $S_A$ via the explicit chest stack, and to $S_B$ by building the chain of jars (i.e., the outermost jar Rico is carrying contains machinery corresponding to the top symbol of $S_B$ as well as a second jar which itself contains machinery corresponding to the next symbol of $S_B$, etc.). During the checking phase, Rico will need to traverse back up the history of chests and prove that each room corresponding to a symbol at the top of $S_A$ matches the symbol corresponding to the room contained in the outermost jar, i.e. at the top of $S_B$, and ``popping'' both off their stacks. Rico can reach the crystal only if they reach the starting room (thereby having emptied $S_A$) when $S_B$ is also exactly empty, at which point Rico will be carrying the initial block rather than a jar.

\subsection{Gadgets}
\label{sec:gadgets}

In this section, we will enumerate a collection of gadgets which will be used in the overall construction. A gadget is a template for a section of a level with specific properties. We first describe the \ow, \poh, and \oto{} gadgets which are simple and useful subcomponents we will use repeatedly. Section~\ref{sec:pv} describes the \pv gadget which has one entrance which can only be traversed if another entrance has previously been traversed. It is the main component in our ability to record state in our construction. Finally, Section~\ref{sec:cr} has multiple parts which detail how dominoes in the PCP are constructed using separate gadgets for the choice of domino placement (Section~\ref{sec:dc}), the top symbol (Section~\ref{sec:aij}), and the bottom symbols (Section~\ref{sec:bij}).

\subsubsection{\ow{}}
\label{sec:ow}
A \ow{} gadget allows Rico to pass from one side of the gadget to another, but not back in the opposite direction. Our \ow{} gadgets have the additional property that Rico is not able to throw an object through one without traversing it themself.

The traversability requirement can be trivially implemented with two ``stair steps'', each two blocks high, followed by a four block drop. Rico can only jump at most 3 blocks, so after jumping down from the ledge, they cannot get back up. \ow{}s can also be implemented with a ledge: Rico can jump up onto the ledge but then is unable to get back down. We use both implementations, governed by desired layout.

The latter ledge implementation trivially satisfies the thrown object requirement, since objects always pass through ledges. The stair step implementation requires one extra feature, which is to make sure the floor below the four block cliff is a ledge, so that any object dropped over the edge will fall through the ledge and become inaccessible, either by getting trapped in an unreachable pit, or by falling off the bottom of the screen, depending on placement. Further, we add a multi-block `stalactite' over the ledge to ensure that any item thrown from above the stairs will hit this wall and fall below the ledge.

The ledge \ow{} can be seen on the far left sides of Figures~\ref{fig:dc} and \ref{fig:aij}. The stair implementation can be seen once in the bottom of Figure~\ref{fig:dc} and in triplet in the bottom of Figure~\ref{fig:aij}.

\subsubsection{\poh{}}
\label{sec:poh}
The \poh{} gadget (H) is a simple gadget which is traversable if and only if Rico is carrying an object. It has the additional important property that the held object must also traverse the gadget, and cannot be left at the entry side of the gadget for later retrieval. See Figure~\ref{fig:poh}. It makes use of the fact that while carrying something Rico's jump height is lower. If Rico jumps three blocks high, which is unavoidable while not carrying an object, they will get stuck in the enclosed area at the top of the gadget. However, if Rico is carrying an object, they will jump only two blocks high and land on the lower edge, and have space to walk out of the gadget to the right. The pit at the bottom of the gadget prevents Rico from dropping the held object back down and leaving it behind (accessibly), as it will get stuck in the pit.

\subsubsection{\oto}
\label{sec:oto}
\begin{figure}
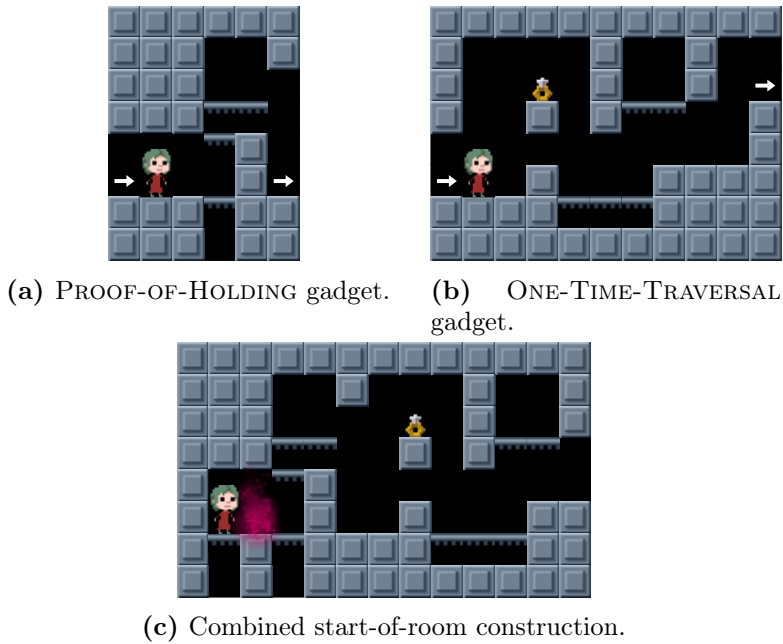

\centering
\subcaptionbox{\poh{} gadget. \label{fig:poh}}[.35\linewidth]{\input{figs/small_examples_proof_of_holding.svg_tex}}
\subcaptionbox{\oto{} gadget. \label{fig:oto}}{\input{figs/small_examples_one_use_oneway.svg_tex}}
\subcaptionbox{Combined start-of-room construction.\label{fig:combined}}[.5\linewidth]{\input{figs/small_examples_combined_poh_oto.svg_tex}}

\caption{The \poh{} and \oto{} gadgets. These exclusively appear together and at the start of most rooms, so we will always use the combined version shown in~\textbf{(c)} for compactness. Note that Rico cannot jump from the ledge on the left directly to the ring, as they will necessarily bump their head on the `stalactite' block and fall, even while holding an object.}
\label{fig:pohoto}
\end{figure}

The \oto{} (1O) is what it says on the tin. Rico can traverse it one time in one direction, after which it cannot be traversed in that direction again. See Figure~\ref{fig:oto}. It is implemented by forcing Rico to be holding an object (we use a ring so as not to bestow any other abilities) in order to jump up a small step, without irreversibly getting stuck on the ledge 3 blocks up. If Rico is not holding the ring while jumping up the step, getting stuck on the ledge is unavoidable. The gap below the ring's tile is to allow Rico to throw a held object over to the other side of the gadget to be retrieved after traversal (recall that Rico, being two tiles high, cannot fit through that gap). The reason the gadget is one-time use is because (1) once the ring is removed it is impossible to get it or any other object up to the tile where the ring is initially and (2) the gadget is only possible to traverse from left to right if there is an object present precisely on that tile.

It is easy to see (1) by recalling that Rico cannot jump to a height of 3 blocks while holding an object, nor is there anyway to throw an object upwards with any horizontal velocity, so Rico can neither carry nor throw an object up to the ring's starting tile. Given (1), (2) becomes clear because there is no way Rico can be carrying an object while standing on the lower ledge \emph{except} by  grabbing one off the ring's starting tile.

For notational convenience, and because we always want to force Rico to prove that the jar is never dropped, we will always combine \oto{} with \poh{} gadgets, to get a gadget which Rico can traverse if and only if they are carrying something and even then at most once. We denote this combined gadget by $1O+H$, or $\nrightarrow$.

\subsubsection{\pv{} Gadget}
\label{sec:pv}
The primary driving gadget behind much of our construction is what we call a \pv{} (PV) gadget. The basic idea is that the gadget primarily consists of a single room which contains a green glowing block which can be in one of two states: set or unset. If Rico is able to visit a \prv{} chest, $P$, they can put the gadget in the set state. If Rico visits a \ver{} chest, $V$, they will be able to retrieve a key from $V$ if and only if the gadget is Set, and in doing so must return it to the Unset state. In this way, retrieving the key from $V$ verifies that $P$ was visited. Note that a \ver{} chest is always followed by a locked door. Unless otherwise stated, PV gadgets are initially unset. We note that this is a minor variation on the ``Self-closing Door'' gadget~\cite{anipspace} in the framework of~\cite{demaine2018computational}.

A major use case for PV gadgets is to force Rico to prove that a room is being entered when and from where it is intended to be. To enforce this, for most rooms $R$, the first element encountered will be a $V_R$ gadget corresponding to that particular room, which will be set only if Rico is coming into that room immediately after visiting a corresponding $P_R$ gadget in the previous room. Correspondingly, whenever we intend Rico to continue on to $R$ in the intended call stack, we will precede the chest containing $R$ with a $P_R$ gadget followed by a one-way.

The crux of the gadget is a stateful \emph{memory room} $M_x$ shared by a \pv{} pair $P_x$ and $V_x$, shown in Figure~\ref{fig:pv-m}. If the green block is in the pit, the gadget is Unset, for Rico cannot retrieve the key (or indeed, leave the gadget at all once falling down the cliff next to the entry). If the block is not in the pit, Rico can jump on it to retrieve the other block up on the ledge, and put both of them in the pit which is enough to get up to the key and back around to the exit. Of course, after doing so the green block is in the pit and the gadget is unset again.

\begin{figure}
\centering
\def\svgwidth{0.45\linewidth}
\subcaptionbox{Unset state. \label{fig:unset}}{\input{figs/undecidable_Kx.svg_tex}}
\qquad
\subcaptionbox{Set state. \label{fig:set}}{\input{figs/undecidable_KAx0.svg_tex}}
\caption{The \mem{} component of the \pv{} gadget.}
\label{fig:pv-m}
\end{figure}

The \prv{} gadget simply gives Rico a block to take into the corresponding \ver{} chest with which they can retrieve the green block from the pit (by using both the extra block and the block up on the ledge), shown in Figure~\ref{fig:prv}. At no point can anything but the block enter the \mem{} room, due to the 3 block high barrier which Rico cannot carry any object over. Similarly, no object but the key may exit the \mem{} gadget, and bringing the key out of the \mem{} chest while in the \prv{} chest is clearly not useful due to the same barrier preventing it going anywhere else.

\begin{figure}
\centering
\subcaptionbox{\prv{} gadget $P_x$. \label{fig:prv}}{\input{figs/small_examples_px_small.svg_tex}}
\qquad
\subcaptionbox{\ver{} gadget $V_x$. \label{fig:ver}}{\input{figs/small_examples_vx_small.svg_tex}}
\caption{The \prv{} and \ver{} gadgets.}
\label{fig:pv}
\end{figure}

The \ver{} gadget, shown in Figure~\ref{fig:ver}, is intended to allow retrieval of a single key if the corresponding \mem{} room is set. It is a bit more complicated in order to prevent cheating. The intended usage is to jump into the \mem{} room and retrieve a key, then throw that key against the door on the left to allow access to the \emph{other} key, which can then be brought out of the \ver{} chest.

It is important to maintain the invariant that no object (other than a block from the corresponding \prv{} gadget) can enter the \mem{} chest, because if Rico could bring in, say, another chest, they could then move the green glowing block from the \mem{} chest into some other room, and this could subsequently result in Bad Things. First, note that the \mem{} chest only appears in the \prv{} and \ver{} gadgets. For the \prv{} gadget, the 3 tile high barrier ensures that nothing can be brought from the entrance of the gadget to the \mem{} chest. For the \ver{} gadget, we again use a 3 tile high barrier which nothing can be carried over. However, in order to allow the \mem{} chest state to interact with the rest of the gadget, we require a gap which the key retrieved from the \mem{} room can be thrown through, opening the door on the left. The important observation is that no object can be usefully thrown through the gap \emph{except} a key going from right to left hitting and opening the door. Any other object will hit a wall or the door and land inaccessibly in the pit under one of the ledges. Thus, again, no object can enter the \mem{} chest, and no object can leave the \mem{} chest except a key, and therefore no object can leave the \ver{} chest except the key behind the door. Finally, we need to ensure that the \mem{} chest itself cannot exit the \prv{} or \ver{} gadgets. Once again, the 3 tile high barriers and the ledges also prevent this.

Note that Rico could bring in a key from outside to open the door with, but all this would achieve is replacing the old key with a new key, so doing so cannot be eminently useful.

\subsubsection{\gl{}}
\label{sec:gl}
We will also use a variant of the \pv{} gadget semantics which we will call a \gl{} (GL). The \gl{} can be `set' just once, and subsequently used to retrieve a key any number of times. The \gl{} gadget will only be used to allow Rico to transition between the pushing phase and the checking phase, and subsequently verify that the transition was made. The \gl{} is just a room with a green glowing locked door with a key behind it (see Figure~\ref{fig:gl}). As long as the door is locked, the key is irretrievable, but once Rico is given a key to take into even one chest containing the \gl{} room, they can permanently unlock the door and allow all the other chests containing this room to dispense keys.

\begin{figure}
\centering
\input{figs/small_examples_gl_small.svg_tex}
\caption{The \gl{} gadget.}
\label{fig:gl}
\end{figure}

\subsection{Component rooms} \label{sec:cr}
In this section, we describe the macro-level component rooms of the construction. Many of them are fairly simple to implement in a way that allows Rico to act in the intended way, but have many additional complications which prevent Rico from cheating, usually in the form of \pv{} gadgets separated by \ow s.

For every following gadget, there is always a one-way path from any terminus back to the exit, since we will need to pop all the rooms off the call stack later.

\subsubsection{\dc{}}
\label{sec:dc}
The \dc{} gadget is constructed with a series of rooms, $DC_{ij}$, which are structured in a binary tree. In each room, shown in Figure~\ref{fig:dc}, Rico can choose which of two paths to take, and each leaf of the tree will contain two chests each corresponding to choosing a particular domino to place (and one additional chest for choosing to stop)\footnote{The playable level provided at~\cite{xls2lua} provides a single $DC$ room with $3$ choices for simplicity. The shortest solution path, after being forced to place $D_0$ first, chooses the middle chest corresponding to $D_1$ upon first visiting the $DC$ room, and the final chest corresponding to \textsc{Stop} on the second and final visit.}. The binary tree or a similar structure is necessary to ensure that the level specification contains only a bounded number of objects per room, which is a real requirement of levels in the game, preventing us from simply packing all domino choice chests into a single room. $DC_{ij}$ is the $j$th possible room Rico can be in at depth $i$ in the tree. 

\begin{figure}
\centering
\input{figs/small_examples_domino_choice.svg_tex}
\caption{The \dc{} rooms.}
\label{fig:dc}
\end{figure}

In order to prevent Rico from cheating later on when popping the call stack, each choice room also has an initial \ver{} chest $V_{DC_{i}}$, which for simplicity is the same for all $DC$ chests at a particular depth $i$, and which verifies that Rico came from an intended place (i.e. a $DC$ chest at depth $i-1$). Correspondingly, each $DC_{ij}$ has a \prv{} chest, $P_{DC_{i+1}}$, followed by a \ow{} preceding chest selection, proving that Rico really is allowed to go into one of the subsequent $DC$ chests now. 

The leaf nodes of the binary tree structure contain chests pointing to rooms $A_{i0}$ for each Domino $i$ (see below). Similarly, the \prv{} chest in the leaf nodes will correspond to $P_{A_{x0}}$, which is a special \prv{} chest corresponding to all $A_{i0}$ rooms simultaneously (see Section~\ref{sec:aij}). If the top half $A$ string of Domino $i$ happens to be empty, then $A_{i0}$ will be replaced with $B_{i0}^{(C)}$. These \pv{} pairs are important because when Rico later pops out of, say, $DC_{i+1,j'}$ when back-traversing the call stack, they cannot (usefully) go back into that chest because the first element will be an untraversable $V_{DC_{i+1}}$ chest, and Rico cannot back-track to the corresponding $P$ chest because it is behind a one-way.


Whichever chest Rico chooses, when they eventually pop back out of it, we will require that they have switched to the checking phase in order to progress, and so we place one final \gl{} gadget after a \ow{} before allowing traversal back to the exit.



\subsubsection{\texorpdfstring{$A_{ij}$}{A\_ij} rooms}
\label{sec:aij}

\begin{figure}
\centering
\input{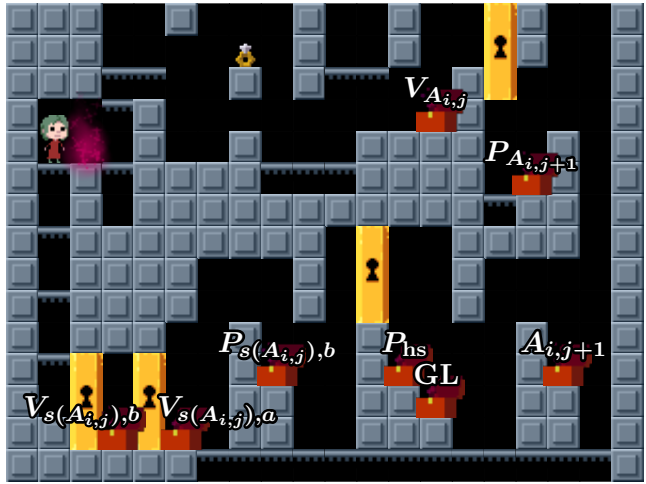}
\caption{The $A_{ij}$ rooms.}
\label{fig:aij}
\end{figure}

Consider the $j$th symbol in the top half of the $i$th domino. We uniquely identify that location by $A_{ij}$ and the (nonunique) symbol by $s(A_{ij})$. Similarly, we label the bottom half locations $B_{ij}$ corresponding to symbol $s(B_{ij})$. Each location $A_{ij}$ has a corresponding room, also labeled $A_{ij}$, shown in Figure~\ref{fig:aij}.

\paragraph{Pushing.}
During the pushing phase, Rico will do the following. Upon entering, Rico must interact with several elements, each separated from the next by a one-way:

\begin{enumerate}
\item Traverse a \poh{} gadget
\item Traverse a  $V_{A_{ij}}$ gadget
\item Traverse a $P_{A_{i,j+1}}$ gadget
\item Enter a chest leading to the next symbol room, $A_{i,{j+1}}$.\label{aij:end}
\end{enumerate}

Note that if $j = 0$, then $V_{A_{ij}}$ will be replaced with $V_{A_{x0}}$ independent of $i$ as described in Section~\ref{sec:dc}. If $A_{ij}$ is the last symbol in the top half string for this domino (i.e. Domino $i$ has exactly $j$ top half symbols), $A_{i,j+1}$ and $P_{A_{i,j+1}}$ will be replaced with $B_{i0}$ and $P_{B_{i0}}$ respectively (see Section~\ref{sec:bij}). If the bottom string of Domino $i$ is empty, the replacements will instead be the first \dc{} room, $DC_{00}$, and $P_{DC_{00}}$, respectively. This is as far as Rico will go during the Domino Selection phase.

\paragraph{Popping.}
Of course, that is not the end of the room, for when Rico jumps back out of the $A_{i,j+1}$ chest they entered in step~\ref{aij:end} above during the checking phase, this is the point where they will need to pop a symbol from $S_B$ and prove that it is equal to $a_{ij}$. Note that Rico will not be able to usefully go back into the $A_{i,{j+1}}$ chest they just jumped out of, since there will immediately be an untraversable $V_{A_{i,{j+1}}}$ gadget.

During the checking phase, Rico must take the following steps, comprising a 2-way handshake to prove that $S(A_{ij})$ and $b_{i'j'}$ are equal to each other. Again, all elements are separated by \ow{}s  (besides entering and exiting held jars, of course).

\begin{enumerate}
\item Traverse the \gl{} gadget to prove the checking phase has been entered and a ``handshake'' chest \phs{}, allowed in either order to save space. The \phs{} gadget will prove that the handshake has been appropriately initiated (see below).
\item Traverse a $P_{s(A_{ij}),b}$ gadget corresponding to the ``bottom half $b$ version'' of $s(A_{ij})$. \label{aij:firstp}
\item Delve into the jar they should be carrying, hopefully containing $B_{ij}^{(J)}$ with  $s(A_{ij}) = s(B_{ij})$. 
\item Inside the jar (see Section~\ref{sec:bij} for details), traverse a $V_{s(B_{ij}),b}$ gadget, only possible if $s(A_{ij}) = s(B_{ij})$.
\item Traverse a $P_{s(B_{ij}), a}$ gadget corresponding to the ``top half $a$ version'' of $s(A_{ij})$.
\item Traverse a \vhs{} chest.
\item Exit the jar, thereby destroying the $B_{ij}^{(J)}$ instance, effectively popping $S_B$. 
\item Traverse the $P_{s(A_{ij}),b}$ from step~\ref{aij:firstp} \emph{again} (see below for an explanation). \label{aij:secondp}
\item Traverse a $V_{s(A_{ij}),a}$ gadget, again only possible if $s(A_{ij}) = s(B_{ij})$.
\item Traverse an instance of $V_{s(A_{ij}),b}$ to re-unset it from step~\ref{aij:secondp}. \label{aij:extrav}
\end{enumerate}

The \phs{}-\vhs{} handshake pair is necessary to disallow popping multiple instances of $s(A_{ij})$ off of $S_B$. Without it, after step~\ref{aij:secondp}, Rico could enter the new top jar, and traverse it successfully, contingent on the symbol it corresponds to being equal to $s(A_{ij})$. However, the \vhs{} gadget prevents this, since it will become unset after having traversed the previous intended jar. 

Steps~\ref{aij:secondp} and~\ref{aij:extrav} are necessary because there is no way to prevent Rico from traversing the $P_{s(A_{ij}),b}$  an extra time after exiting the jar, which could allow future unintended traversals of a $V_{s(A_{ij}),b}$ somewhere else. Thus, we must simply assume that Rico will traverse the $P_{s(A_{ij}),b}$ chest again and then force them to unset the $V_{s(A_{ij}),b}$ in step~\ref{aij:extrav}.



\subsubsection{\texorpdfstring{$B_{ij}$}{B\_ij} rooms}
\label{sec:bij}

\begin{figure}
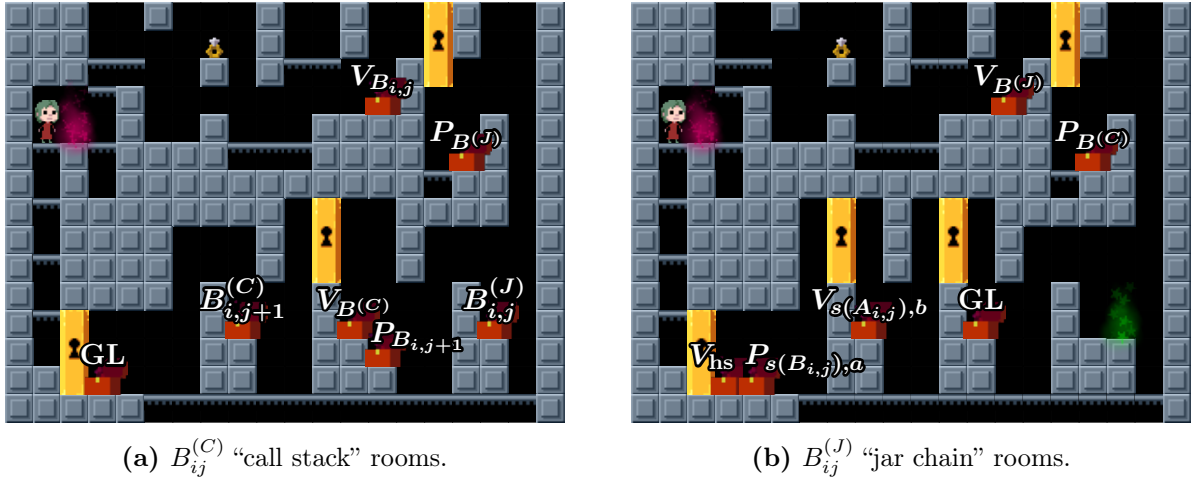

\centering
\def\svgwidth{0.45\linewidth}
\subcaptionbox{$B_{ij}^{(C)}$ ``call stack'' rooms.}{\input{figs/undecidable_BC10.svg_tex}}
\qquad
\subcaptionbox{$B_{ij}^{(J)}$ ``jar chain'' rooms.}{\input{figs/undecidable_BJ10.svg_tex}}
\caption{The $B_{ij}$ rooms.}
\label{fig:bij}
\end{figure}

The $B_{ij}$  rooms are a bit more complicated (yes, really), shown in Figure~\ref{fig:bij}. Recall that Rico is always carrying around a jar containing the chain of jars representing $S_B$, with a prize (block) at the bottom. We need to force Rico to add more rooms to the jar chain. For each symbol $b_{ij}$, we will have two corresponding rooms: one room which will go on our call stack $B_{ij}^{(C)}$ which forces us to place the other, $B_{ij}^{(J)}$, in our jar chain. 

\paragraph{Pushing.}
During the pushing phase, the intended traversal is as follows (each element separated from the next by a one-way):

\def \pj {\ensuremath{P_{B^{(J)}}}}
\def \pc {\ensuremath{P_{B^{(C)}}}}
\def \vj {\ensuremath{V_{B^{(J)}}}}
\def \vc {\ensuremath{V_{B^{(C)}}}}

\begin{enumerate}
\item Traverse a \poh{} gadget.
\item Traverse $V_{B_{ij}}$.
\item Traverse \pj{}.
\item Enter $B_{ij}^{(J)}$ carrying the current top jar (or, initially, the block).
\item Traverse a \poh{} gadget, to prove Rico brought the jar chain with.
\item Traverse \vj{} (this prevents Rico from trying to push $s(B_{ij})$ onto $S_B$ more than once).
\item Traverse \pc{} (allowing Rico to prove that this symbol really was pushed, and not skipped).
\item Put down the jar and exit $B_{ij}^{(J)}$ via green flame, creating a new jar with the chain extended by one. The jar must be put down here to allow use of the green flame (recall Section~\ref{sec:jars}).
\item Traverse \vc{}, which is possible only if Rico really did push the new symbol.
\item Traverse $P_{B_{i,j+1}}$.
\item Continue on to $B_{i,j+1}$.
\end{enumerate}

As with the $A_{i,j+1}$ chests in Section~\ref{sec:aij}, if $B_{ij}$ corresponds to the last $B$ symbol of domino $i$, $B_{i,j+1}$ and $P_{B_{i,j+1}}$ will be replaced with $DC_{00}$ and $P_{DC_{00}}$ respectively.

\paragraph{Popping.}
During the checking phase, the $B_{ij}^{(C)}$ are very simple. Rico does not need to prove anything while popping out of $B_{ij}^{(C)}$, and Rico cannot jump back into the $B_{i,{j+1}}$ chest they came from since they will be met by an untraversable $V_{B_{i,{j+1}}}$. Thus, Rico just traverses a \gl{} gadget to prove that they really are in the checking phase, and then loops back up to the exit and pops up.

\begin{figure}
\centering
\input{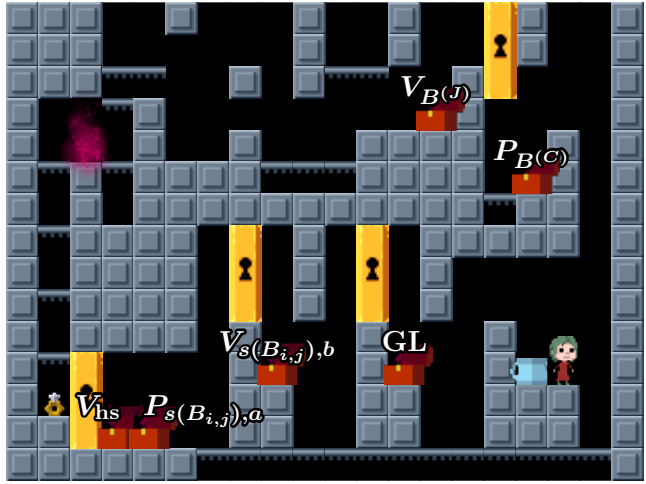}
\caption{The $B_{ij}^{(J)}$  rooms while they are in the jar chain. Note, relative to the room's initial state: (1) the \oto{} gadget has been traversed, so the ring has been moved (e.g. by being thrown against the far right wall); (2) Rico has used the green flame, so it is gone (but Rico appears where it used to be); (3) Rico has dropped the jar containing the next $B_{ij}^{(J)}$ room in the chain next to the re-entry point, for retrieval.}
\label{fig:jarchain}
\end{figure}

The $B_{ij}^{(J)}$ rooms, on the other hand, contain the two-way handshake machinery described in section~\ref{sec:aij}. Figure~\ref{fig:jarchain} depicts the (intended) intermediate state of the $B_{ij}^{(J)}$ rooms at the point when Rico re-enters the corresponding jar. To reiterate, Rico is intended to go into the top level jar containing $B_{ij}^{(J)}$ after traversing \phs{} and $P_{s(B_{ij}),b}$ in the corresponding $A_{i'j'}$ room (where $s(B_{ij}) = s(A_{i'j'})$ if the solution is correct), but before traversing $V_{s(A_{i'j'})}$. Given this, Rico can:

\begin{enumerate}
\item Jump into the jar containing $B_{ij}^{(J)}$.
\item Traverse a \gl{} gadget to prove that they are in the checking phase.
\item Traverse $V_{s(B_{ij}),b}$.
\item Traverse $P_{s(B_{ij}),a}$.
\item Traverse \vhs{}.
\item Carry the jar containing the rest of the chain back to the exit, returning to the $A_{i'j'}$ room with the new jar.
\end{enumerate}




\subsubsection{Top level (start room)}
\label{sec:start}
As described in Section~\ref{sec:highlevel}, Rico initially spawns in a room with a block and an $A_{00}$ chest, a \gl{} chest behind a \ow{}, and the crystal 4 blocks up behind a locked door, shown in Figure~\ref{fig:start}. 
 
The $A_{00}$ room has a $V_{A_{x0}}$ chest (see Section~\ref{sec:aij}) which is specially initialized to Set (unlike every other $V$ chest), since we enforce that domino $D_0$ must be chosen first (this way we prevent winning by placing no dominoes). Rico can take the block into $A_{00}$, and in fact, must do so in order to pass the first \poh{} gadget.  This very special block will be in place of the final jar at the very end of the chain. This block is so special because it is the \emph{only} block in the entire level which is not contained in a \pv{} gadget. Furthermore, because it exists in the starting room, which is not contained in any chests, the block can never be duplicated by any means. Rico could use this block to cheat and falsely traverse a $V$ gadget of their choice, but this would permanently lose the block and Rico will never be able to get another block outside of $PV$ gadgets and thus will never be able to reach the goal crystal. Rico probably should not do that. This property is also why Rico should never leave behind the chain of jars containing the block; if the jar chain is ever lost, the goal will be unreachable. The block cannot help Rico do anything else (except reach the goal of course!), thus, it is of no more use than a jar until the very end.

If Rico tries to traverse the \ow{} before opening the \gl{} and returning to this room as in an intended solution, they will necessarily get stuck either without the block (necessary to reach the crystal) or without the $A_{00}$ chest (necessary to have any way to open the \gl{} chest) and be unable to complete a solution. 

If Rico pops back up the call stack to the very first room (so back out of the first $A_{00}$) while carrying the block (having traversed precisely all of the jar chain, Rico can open the locked door using the \gl{} chest, jump on the block and reach the crystal! If Rico reaches this point but is still carrying a jar and not the block, the $V$ gadgets in the jar will be untraversable and Rico will not be able to ever reach the block. If Rico reaches the block in the jar chain without popping the call stack all the way back to the starting room, the $V$ gadgets in the remaining $A_{ij}$ rooms will be untraversable (at least not without losing the block) and Rico cannot get back to the goal.

\subsection{Changing phases}
\label{sec:stop}
\begin{figure}
\centering
\input{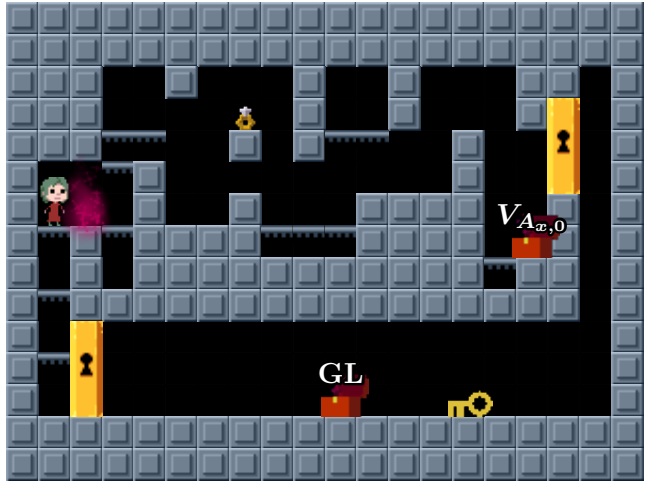}
\caption{The \textsc{Stop} room which allows Rico to switch from the Pushing phase to the Checking phase.}
\label{fig:stop}
\end{figure}

 When Rico wants to switch from the pushing phase to the checking phase, they must choose the \textsc{stop} chest in the \dc{} room. The \textsc{stop} chest is a comparatively simple room, shown in Figure~\ref{fig:stop}. It first contains a \poh{} gadget, followed by a $V_{A_{x0}}$ \ver{} chest (see Section~\ref{sec:dc}) followed by a \gl{} chest\dots\ along with a key! Rico can thus permanently unlock the \gl{} chest and allow all its instances to produce keys.

At this point, all Rico can do is leave the room and return to the \dc{} room. From now on, Rico should never again be able to visit any of the $P_R$ chests which correspond to $V_R$ chests at the start of each room, so Rico should never be able to generate and traverse a fresh instance of any of the DC, $A_{ij}$, or $B_{ij}$ rooms. Popping mechanisms for each room type are described above.

\subsection{Victory} If Rico has managed to pop all the way back up the call stack \emph{and} clear out the chain of jars so that they are again directly carrying the block, they can traverse the open \gl{}, obtaining a key to open the door, and using the block to jump and reach the crystal!

\subsection{Reduction accounting}
How many rooms are there? Suppose we have $k$ possible dominoes, and the sum of the length of the strings on the top half of dominoes is $t_A$ and the sum of the lengths of the strings on the bottom half of dominoes is $t_B$. Suppose the number of distinct symbols is $s$.

\begin{itemize}
\item $k-1$ $DC$ rooms (since the $DC$ rooms form a binary tree structure with $k/2$ leaf nodes)
\item $3\lceil \log_2 k \rceil$ rooms corresponding to DC \pv{} gadgets (\prv{}, \ver{}, and \mem{})
\item $t_A$ $A_{i,j}$ rooms 
\item $3$ rooms for \pv{} gadgets corresponding to $A_{x0}$
\item $3(t_A - k)$ rooms for \pv{} gadgets corresponding to $A_{ij}$ rooms, other than $A_{x0}$
\item $2t_B$ $B_{i,j}$ rooms (of $(C)$ and $(J)$ varieties)
\item $3t_B$ rooms for \pv{} gadgets corresponding to $B_{ij}^{(C)}$ rooms
\item 6 $P_{B^{(C)}}$, $P_{B^{(J)}}$, and corresponding \ver{} and \mem{} rooms
\item $6s$ rooms for \pv{} gadgets $P_{s,a}$, $P_{s,b}$, etc.
\item 3 \phs{} and \vhs{} and $M_{hs}$ rooms
\item 1 starting room
\item 1 \textsc{Stop} room
\item 1 \gl{} room
\end{itemize}

The resulting grand total is $ 4t_A + 5t_B - 2k + 3\lceil \log_2 k \rceil +6s +14$, which is notably linear in the PCP instance size. This confirms that the Recursed level specification has size linear in the size of a Turing Machine simulated by a PCP instance.

Correspondingly, the length of the (shortest) solution path is $L \log k$ where $L$ is the length (number of symbols) of the PCP solution string and, as above, $k$ is the number of possible dominoes. In particular, Rico must traverse exactly one $A_{ij}, B_{ij}^{(C)},$ and $B_{ij}^{(J)}$ room for each symbol in the solution string. The $DC$ rooms introduce the $\log k$ term, since Rico must traverse a $\log k$ length path through the binary tree of $DC$ rooms in order to select a domino to place at each step. Each domino placed must contribute at least one symbol to at least one of the top half string or the bottom half string, so at most $2L$ dominoes get placed. Each of the chests in the aforementioned room requires a constant amount of time spent traversing their component gadgets (mostly \pv{} rooms), resulting in a total of $L\log k$ as claimed. 

\section{Open Problems}\label{sec:open}

Although we achieve a tight result of RE-completeness, we could still ask about the complexity of Recursed with a subset of the puzzle mechanics. We propose two conjectures and two open problem relating to subsets of Recursed mechanics:

\begin{conjecture}[Cauldrons but no Jars] We conjecture that Recursed with Cauldrons, but no Jars, is still undecidable.
\end{conjecture}

This conjecture seems very likely because Cauldrons (which are intended to intuitively represent multi-threading) allow Rico to jump between different ``worlds'' (up to $4$, represented visually by background color) which \emph{each have their own chest history}. Thus, it should not be difficult to build a reduction similar to ours which makes use of multiple stacks to simulate PCP or 2-stack Push Down Automata, both of which are undecidable.

\begin{conjecture}[No Cauldrons or Jars]  We conjecture that Recursed without Cauldrons or Jars can be simulated by a Push-Down Automata, and is therefore undecidable.
\end{conjecture}

The main difficulty with proving this conjecture is that during a solution, rooms can contain an unbounded number of objects (blocks, keys, or chests), and such state can not be trivially stored in either the automata head, or on the stack. However, we conjecture that after some bounded point, more objects of a given type cannot help towards a solution, and can therefore be forgotten. However, this seems difficult to prove.

\begin{open}[Jars or Cauldrons but no Green Glow]  What is the complexity of Recursed with Jars or Cauldrons or both, but without green glowing objects?
\end{open}

Green glowing objects are not required to build some form of two or more stateful stacks with either Jars or Cauldrons, but it seems very difficult to construct reductions without them. It's possible that there simply is not enough interaction amongst the limited set of objects in Recursed for this problem class to be undecidable, but it seems quite difficult to rule out.

\begin{open}
What is the complexity of Recursed restricted to a polynomial-length ``room stack'' (analogous to call stack)?
\end{open}
 This problem is naturally in NPSPACE${}={}$PSPACE, but is it PSPACE-complete?  This question likely needs a different approach, as our reduction is focused on time simulation and not on multiple uses of gadgets.

\section*{Acknowledgments}

This work was initiated during the 33rd Bellairs Winter Workshop on
Computational Geometry, co-organized by Erik Demaine and Godfried Toussaint
in March 2018 in Holetown, Barbados.
We thank the other participants --- in particular, Robert Hearn ---
for related discussions and providing an inspiring atmosphere.
We thank Edison Y. He for his helpful comments on earlier drafts of this paper.
Figures were generated using SVG Tiler \cite{svgtiler}.

\bibliographystyle{alpha}
\bibliography{biblio}

\end{document}